\documentclass[sigconf]{acmart}

\usepackage{booktabs} % For formal tables

\setcopyright{rightsretained}
\usepackage{amssymb}
\usepackage{graphicx}

\usepackage{url}
\usepackage{epstopdf}

\usepackage{xcolor}
\usepackage{soul} 

\begin{document}
\sloppy

% first the title is needed
\title{How To Extract Fashion Trends From Social Media?}
\subtitle{A Robust Object Detector With Support For Unsupervised Learning}

\author{Vijay Gabale}
\affiliation{%
  \institution{Infilect Technologies Pvt. Ltd.}
  \city{Bangalore}
  \state{India}
}
\email{vijay@infilect.com}

\author{Anand Prabhu Subramanian}
\affiliation{%
  \institution{Infilect Technologies Pvt. Ltd.}
  \city{Bangalore}
  \state{India}
}
\email{anand@infilect.com}

\begin{abstract}
With the proliferation of social media,
fashion inspired from celebrities, reputed designers
as well as fashion influencers has 
shortned the cycle of fashion design and manufacturing.
However, with the explosion of fashion related content
and large number of user generated fashion photos,
it is an arduous task for fashion designers to wade through
social media photos and create a digest
of trending fashion.
Designers do not just wish to have fashion related photos
at one place but seek search functionalities that can let them
search photos with natural language queries such as
`red dress', 'vintage handbags', etc in order to spot the trends.
This necessitates deep parsing of fashion photos on social
media to localize and classify multiple fashion items from a
given fashion photo.
While object detection competitions such as MSCOCO have thousands
of samples for each of the object categories, it is quite difficult
to get large labeled datasets for fast fashion items.
Moreover, state-of-the-art object detectors~\cite{LiuAESRFB16,RedmonCVPR16,FuLRTB17}
do not have any functionality to ingest large amount of unlabeled data
available on social media in order to fine tune object detectors with labeled datasets.
In this work, we show application of a generic object detector~\cite{CDSSD18}, that can 
be pretrained in an unsupervised manner, on 24 categories
from recently released Open Images V4 dataset.
We first train the base architecture of the object detector using
unsupervisd learning on 60K unlabeled photos from 24 categories gathered
from social media, and then subsequently fine tune it on
8.2K labeled photos from Open Images V4 dataset.
On $300 \times 300$ image inputs, 
we achieve 72.7\% mAP on a test dataset of 2.4K photos while
performing 11\% to 17\% better as compared to the state-of-the-art
object detectors. 
We show that this improvement is due to our choice
of architecture that lets us do unsupervised learning and that
performs significantly better in identifying small objects.
\footnote{We are in the process of open sourcing details of labeled datasets
chosen, links to unlabeled datasets, and trained fashion detection models.}

\end{abstract}


\maketitle

\section{Introduction}

\begin{figure*}
\centering
\includegraphics[width=\textwidth]{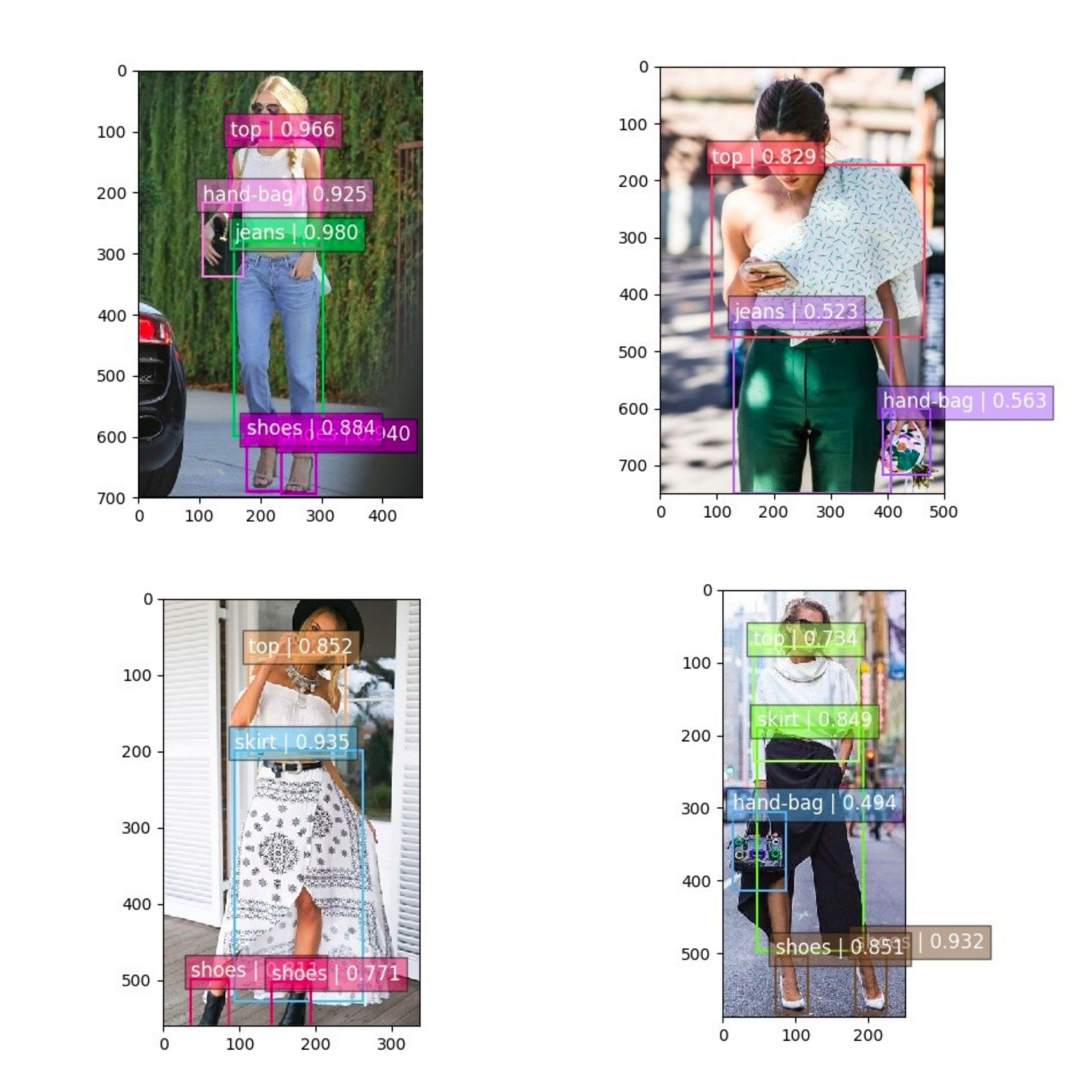}
\caption{Preview of detection outputs using our technique}
\label{fig:ssd-stairnet-bssd}
\end{figure*}

Fashion designers today actively seek inspirations from
social media photos to formulate 
innovative designs before they are put into production.
In last several years,
social media is flooded with fashion inspirations
from celebrities, reputed designers as well as fashion influencers.
However, it has only made the task of manually parsing and extracting intelligence
from these photos arduous for fashion designer.
Fashion designers today wish to have 
an easy to use tool with several search functionalities 
that can let them search
photos with natural language queries 
such as `red dress', `flora handbags', `vintage tees', etc.
The main bottleneck in creating this digest is
the identification of fashion photos and fashion items
inside these photos.
This necessitates deep
parsing of fashion photos on social media to localize and classify
multiple fashion items from a given fashion photo.

Image object detection involves
identifying bounding boxes encapsulating objects 
and classifying each bounding box to recognize
the underlying object category.
Recently
there has been mounting interest in the research
community to detect multiple objects
in an image using Single Shot Detection techniques~\cite{LiuAESRFB16,RedmonCVPR16}.
These techniques effectively combine region proposal
and classification into a single step
by foregoing the candidate box proposal (or region proposal)
module employed by several two-step detection techniques
\cite{GirshickIccv15FastRCNN,GirshickCvpr14RCNN,FasterRCNNNips15,ion16,fpn16}.

While these techniques have made some progress,
they fundamentally lack two key features to
solve problems such as detecting fashion objects
from fashion photos in the wild:
(1) ability to ingest large amount of unlabeled
dataset and (2) ability to maintain detection accuracy
in the absence of large labeled datasets for small
and big object alike.
These two problems are especially relevant to parse
fashion photos on social media since social media
has large amount of unlabeled fashion photos,
i.e., fashion photos for which classes and boxes
are absent.
Moreover, there are no labeled datasets that have several thousands
of labeled photos for each of several tens of fashion
categories.
It is not easy to get such large dataset labeled
owing expertise, cost, and timing constraints.

In this work, we apply a convolution-deconvolution based
object detector to extract fashion objects from fashion photos.
Specifically,
\begin{itemize}
\item We apply an end-to-end trainable convolution-deconvolution based
single shot detection framework to detect multiple objects in an image.
This framework enables unsupervised pretraining of the underlying network.
%in an unsupervised way.
\item Our detection framework enables us to pretrain the model on 60K fashion
photos in the wild. Subsequently, 
we train our technique on 8.2K photos from Open Images V4 dataset
and test it on 2.4K photos.
\item We compare our object detector with state-of-the-art
object detectors. We take object detection models pretrained
on PASCAL VOC dataset and then fine tune them on fashion datasets.
We show that our model performs 11\% to 17\% better. 
\end{itemize}

\section{CDSSD Architecture}

We apply the technique proposed in CDSSD~\cite{CDSSD18}
to detect fashion objects from fashion photos.
In this section, 
we first give a primer on SSD architecture
that is popularly used for object detection.
We then explain why this architecture is
inadequate for us to process social media fashion photos.
We then briefly explain how CDSSD architecture extends SSD
and discuss how we benefit by applying CDSSD architecture
to extract fashion objects from fashion photos
in the wild.

\subsection{SSD}

The SSD network is a convolutional architecture 
that utilizes different layers to predict presence
of multiple objects in an image.
To recognize objects at different scales,
SSD utilizes predictions on different feature maps,
each from a different layer, of a single network.
Instead of processing the image at different sizes,
these feature maps are processed 
by a fixed-size collection of bounding boxes
customized for each layer.
The boxes are applied on each feature map.
Each default box is then evaluated for 
the presence of object class instances.
For feature map $f$ of size $m \times n$ with $p$ channels,
$K$ default-sized bounding boxes are applied on each of $m \times n$ cells.
Subsequently, $C$ filters of size $3 \times 3 \times p$ 
are applied for each cell and for a given bounding box to produce individual scores
to predict each of $C$ classes,
and $4$ additional filters are applied to produce offsets
(center co-ordinate, height, width) to position the box
on the underlying cell in order to encapsulate the object
(as shown in Fig.~\ref{fig:combined1}(c)).
Note that, for a given feature map $f$, 
the default boxes are scaled with a scaling factor
$f^{scale}$ with respect $m$ and $n$ and thus, they
are customized to have different aspect ratios.
Hence, bounding boxes on initial stage feature maps cover a smaller receptive field to identify
objects at a smaller scale, whereas bounding boxes on later stage feature maps
cover larger receptive fields to identify objects with larger scale.
By utilizing predictions 
for all the default boxes with different scales and aspect ratios
from all locations of many feature maps, 
SSD attempts a diverse set of predictions, covering
various input object sizes and shapes.

While SSD technique is useful to detect fashion objects from fashion photos,
it has two main limitations.
\begin{itemize}
\item It does not have any provision to input unlabeled fashion photos, and
train the feature maps in an unsupervised fashion. Thus, although social media
has several hundreds of thousands of fashion photos, these photos can not be
directly applied unless they are labeled with classes and bounding boxes. 
\item It is known to perform poorly to identify small sized objects.
This is especially a concern for fashion photos since object categories such
as earrings or clutches are small in size and appear in all sorts of sizes and shapes.
Furthermore, since tops and tees or such fashion pairs look quite similar,
extracting fashion objects from photos in the wild is necessarily a harder problem
than extracting objects such as planes vs people that are relatively easy to distinguish.
\end{itemize}

\subsection{CDSSD Architecture}

\begin{figure*}[t]
\centering
\includegraphics[width=\textwidth]{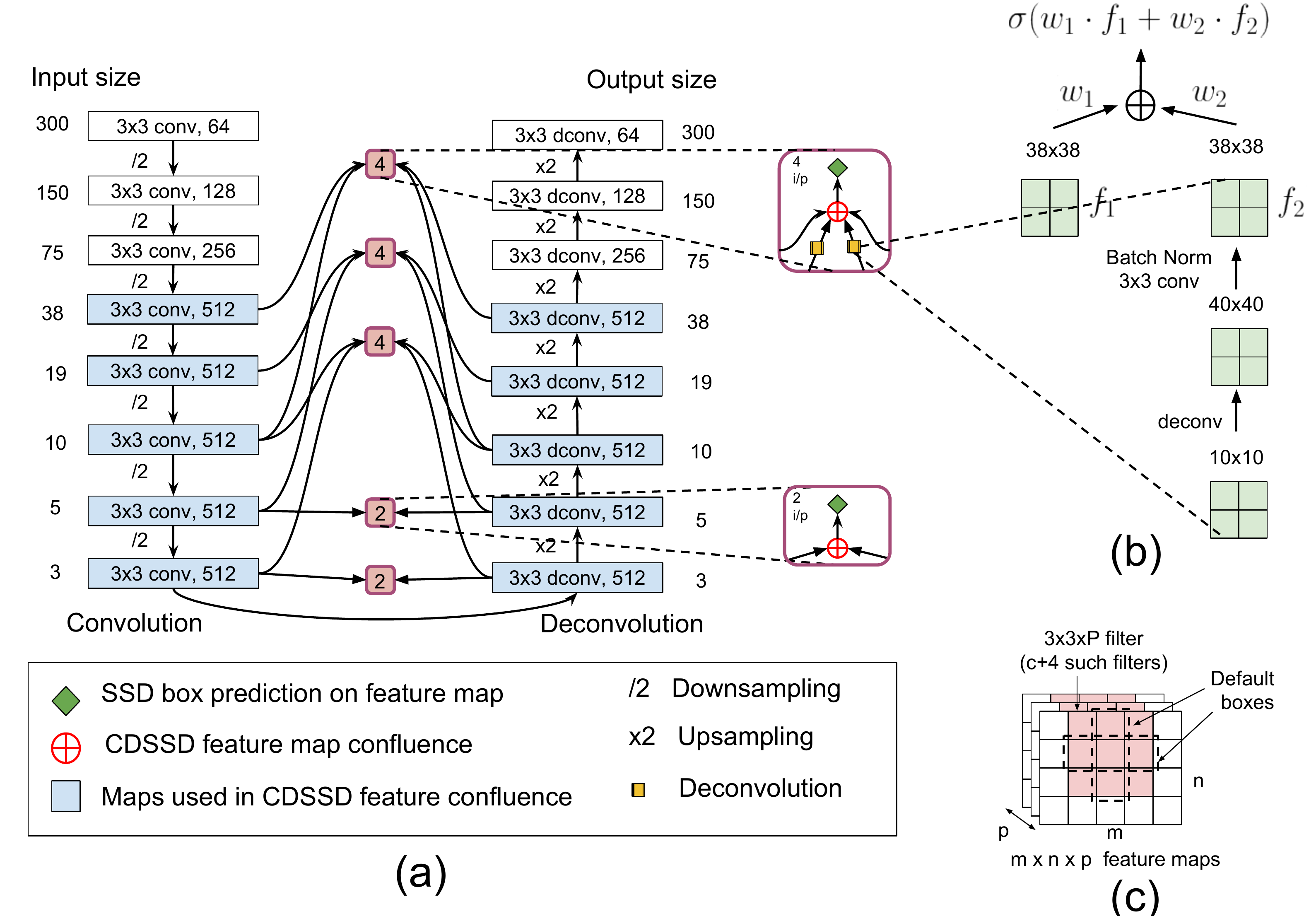}
\caption{CDSSD combines information from convolution and deconvolution feature maps}
\label{fig:combined1}
\end{figure*}

As mentioned in~\cite{CDSSD18},
CDSSD facilitates unsupervised training of 
the underlying network architecture.
For the purpose of this work,
we use ResNet 101 architecture~\cite{resnet101} and construct
a convolution-deconvolution based auto-encoder
(shown in Fig.~\ref{fig:combined1}(a)).
The deconvolution block produces an image of the same dimension as input.
We use an input image of $300 \times 300 \times 3$, with 7 meta-layers of convolution
and pooling and 7 meta-layers of deconvolution with learned upsampling.
Given an image dataset, we first pretrain the architecture 
several thousands of fashion photos in the wild.
After pretraining, we fine tune the same network 
by applying supervised object detection.

It is well-known that feature maps in different layers
of a network have different receptive fields and hence
they learn different set of features; initial layers
especially learn generic features whereas final layers
learn semantically rich features.
\cite{objectdetectorref1,objectdetectorref2} observe
that the initial layers of a deep network lack 
strong semantic information and respond to only
high-level features of an image.
Furthermore, the improvement in acquiring semantic
information across consecutive feature maps is only marginal,
especially in initial layers of a network.
CDSSD exploits these observations
and fuses generic and semantic features to enrich feature maps.
Furthermore, CDSSD combines all the four feature maps of a given
meta layer as shown in Fig.~\ref{fig:combined1}(a).
Similar to SSD, CDSS then applies a set of $K$ default-boxes
and $(C + 4) \times m \times n \times K$ filters on the resulting
feature map to predict detection of objects.
Note that, since 6th and 7th meta-layers have higher reception field
and contain richer semantic information,
they are quite capable of 
detecting bigger size and scale objects~\cite{FuLRTB17}.

This feature especially helps us to locate and classify
small size objects such as earrings, sandals, hats, boots,
wrist-watches, sunglasses from photos in the wild.
Similar to CDSSD,
to compute different aspect ratios for each cell, we take a statistical approach
and compute a cumulative distribution of aspect ratios
of the ground truth boxes in a given dataset.
We then divide the distribution into $B$ bins
and pick the average value of a bin as one of the aspect ratio,
thus resulting in $B$ aspect ratios.
For each $b_i \in B$, 
for a feature map with size
$m \times n$ and scale of $f^{scale}$,
we then set height to be
$m \times b_i \times f^{scale}$ and width to be
$n \times b_i \times f^{scale}$.
With optimized aspect ratios that fit the underlying dataset
and different scales for different layers,
we apply appropriate default boxes at box-pooled locations
in each feature map, covering different object sizes and shapes.

We trained CDSSD architecture on 60K social media photos
in an unsupervised fashion, and 8.2 photos in a supervised
photos. After training, we applied the model on 2.4K photos
for evaluation. Furthermore, we applied the model to more than
500K fashion photos, and extracted fashion objects and monthwise
trends of different fashion styles.

\section{Results}
\label{results}

\begin{table*}
  \centering
  \caption{Comparison of single-shot detection techniques on 2.4K fashion photos,
           trained on 8.2K fashion photos}
  %CDSSD outperforms other state-of-the-art methods while maintaining
  %high speed of detection.} 
\begin{tabular}{c | c c c c c}
method & network & mAP & number of default boxes & fps & lib \\
\hline
YOLOv2\_352~\cite{RedmonCVPR16} & DarkNet-19 & 56.7 & 98 & 81 & DarkNet \\
DSSD321~\cite{FuLRTB17} & ResNet-101 & 63.6 & 43688 & 9.5 & Caffe\\
Stairnet~\cite{stairnet17} & VGGNet & 64.8 & 8732 & 30 & PyTorch\\
CDSSD300 & ResNet-101 & \textbf{72.7} & 1182 & 51 & TF\\
\end{tabular}
  \label{tab:fashionphotosmap}
\end{table*}
%\vspace{-1cm}

\begin{table*}[t]
  \small
  \centering
  \caption{mAP comparison of single-shot detection techniques on fashio dataset}
  %CDSSD results in state-of-the-art performance for several object categories.} 
\begin{tabular}{c | c  c  c   c  c  c  c  c  c  c  c  c }
method & jeans & boot & high-heels & shorts & sandals & briefcase & coat & shirt & brasseire & swimwear & suit & miniskirt\\
\hline
YOLO~\cite{LiuAESRFB16} &  68.5 &  72.2 & 74.8 & 61.9 &  47.6 & 76.7 & 76.8 & 73.5 &  58.1 & 60.0 & 72.4 & 78.9 \\
DSSD321~\cite{FuLRTB17} & 67.6 & 73.3 & 75.4 &  64.6 & 46.8 &  74.7 &  71.5 &  66.9 & 59.5 & 78.3 & 73.2 & 75.4\\  
StairNet~\cite{stairnet17} & 67.0 & 75.4 & 74.2 &  64.2 & 51.3 &  77.6 &  78.0 &  72.0 & 58.9 & 71.8 & 68.4 & 70.2 \\  
CDSSD300 & 77.4 & 73.9 & 78.2 & 89.5 & 84.7 & 80.2 & 80.3 & 78.7 & 73.4 & 69.9 & 76.7 & 83.3\\
\hline
\hline
method & jacket & dress & sun-hat & cowboy-hat & umbrella & glasses & belt & earrings & hand-bag & watch & backpack & suitcase \\
\hline
YOLO~\cite{LiuAESRFB16} & 60.1 & 63.4 & 75.7 & 70.5 & 62.6 & 60.2 & 63.8 & 69.3 &  66.6 & 72.1 & 68.9 & 72.5\\
DSSD321~\cite{FuLRTB17} & 62.3 & 74.5 & 76.2 & 76.6 &  78.1 &  53.3 &  79.6 &  75.7 &  72.2 & 73.9 & 76.3 & 78.4\\
StairNet~\cite{stairnet17} & 65.0 & 69.6 & 56.3 &  74.2 & 62.6 &  73.2 &  75.5 &  61.8 & 66.7 & 52.1 & 70.3 & 71.9\\  
CDSSD300 &  74.5 & 79.4 & 77.1 & 75.2 & 82.6 & 66.4 & 76.1 & 71.4 & 74.7 & 77.9 & 80.8 & 82.5 \\
\hline
\hline
\end{tabular}
  \label{tab:fashiontest}
\end{table*}

\begin{table*}
  \centering
  \caption{mAP at recall greater than 0.7} 
\begin{small}
\begin{tabular}{c | c | c c  c  c | c }
method & data & & & recall &\\
 & & 0.5 & 0.7 & 0.9 & mAP@70\%\\           
\hline
YOLO  & 07+12 & 81.9 & 73.7 & 33.4 & 40.2 \\
SSD & 07+12 & 84.6.3 & 78.7.5 & 33.6 & 45.7 \\
CDSSD & 07+12 & \textbf{98.4} & \textbf{90.1} & \textbf{56.6} & \textbf{62.9} \\
\hline
\end{tabular}
\end{small}
  \label{tab:precision-recall}
\end{table*}

Our experiments are governed to answer 
the following key question:
{\it can we achieve better results in identifying fashion items
from photos in the wild by employing unsupervised learning and confluence of feature maps
from convolution and deconvolution blocks?}
Towards answering this question,
we compare our approach with prior work
on two state-of-the-art techniques: SSD~\cite{LiuAESRFB16} 
and YOLO~\cite{RedmonCVPR16}.
Note that, SSD and YOLO do not employ unsupervised learning
and do not consider confluence contextual and semantic features
from convolution and deconvolution blocks.
We take object detection models for each of these techniques, pretrained
on PASCAL VOC dataset, and then fine tune them on 8.2K labeled photos
from 24 categories.

\subsection{Dataset}

We consider recently released Open Images V4
dataset that has photos and bounding boxes
for the following object categories.

\textbf{Object Categories:}
sandal,
high-heels,
boot,
jeans,
shorts,
swimwear,
brasseire,
shirt,
coat,
suit,
miniskirt,
jacket,
dress,
sun-hat,
cowboy-hat,
umbrella,
glasses,
belt,
earrings,
handbag,
watch,
backpack,
suitcase,
briefcase.

We totally consider 8.2K photos with on an average
600 instances for each object category.
In Addition, we consider an unlabeled dataset of 60K photos
collected from public APIs of social media networks,
Facebook and Instagram.
These photos contain even distribution of the above
categories.

\subsection{Training}
The configuration of our network architecture
is shown in Fig.~\ref{fig:combined1}.
We keep the dropout layers during unsupervised training
and remove them while training for object detection.
We train our models on Azure GPU instances that have
NVIDIA K80 GPUs with 12GB of memory.
We use batch size of 16,
momentum as 0.9 and weight decay 0.0005.
Similar to SSD~\cite{LiuAESRFB16},
we  match  a default  box  to target  ground truth boxes, 
if Jaccard overlap is larger than a threshold (e.g. 0.5).
We compute the target ground truth box for each layer of the network
by scaling it with respect to the feature map and original image sizes.
We minimize the joint localization loss (i.e., smooth L1) 
and confidence loss (i.e., softmax-cross-entropy).  
Note that after the matching step,
most of the default boxes are negatives. 
Hence, to avoid the imbalance between the positive and negative training examples,
we sort the negative boxes using the joint loss for each default box and 
then pick the top ones to maintain a 2:1 negative to positive ratio.
We found 2:1 ratio leads to faster optimization
as compared to the ratio of 3:1 as mentioned in the original SSD paper.

We further make the model robust to different input object sizes and
shapes by invoking extensive augmentation.
Specifically, we sample a patch from a ground truth box so that the
minimum Jaccard overlap with the objects is 0.5, 0.7, or 0.9.
Furthermore, we randomly sample a patch between
[0.5, 1] of the original image size, and the aspect ratio
is between [1, 2].
Also, we randomly flip each patch horizontally with
probability of 0.5, apply different transformations such
as gaussian blur, emboss, edge prominence, random black-out of
20\% of pixels, and color (hue, saturation, contrast) distortions.
We apply $3 \times 3$ box pooling for layer 3 and 4, $2 \times 2$ box pooling
for layer 5, and no box pooling for layer 6 and 7.
We apply non-maximum suppression (NMS) 
to post-process the predictions to get final detection results.

We train the entire network with learning rate at
$10^{-3}$ for 25K batches, and then
with learning rate of $10^{-4}$ for 60K batches to execute
unsupervised pretraining on the underlying train dataset
During object detection training,
we again fine-tune the entire network
with learning rate of $2 \times 10^{-3}$
for 30K iterations, and 60K iterations
with learning rate of $10^{-4}$.

\subsection{Results}
Comparison of MAP for different models
on the test fashion set are shown
in Tab.~\ref{tab:fashionphotosmap}.
Although 2.4K is a smaller dataset
for MAP evaluation, we clearly
get an indication that CDSSD performs
better due to the facility of pretraining
using unsupervised learning.
Note that the same 8.2K labeled dataset
is used to fine-tune all the networks.
Per category results 
are shown in Tab.~\ref{tab:fashiontest}.
These results corroborrate that
that by adding unsupervised pretraining
and confluence of feature maps, CDSSD
consistently outperforms YOLO and SSD
by 11\% to 17\% points
for several object categories.
CDSSD especially shows significant improvement
for small objects such as boots, high-heels, hand-bags.
CDSSD detects majority objects with high confidence
with less localization error and
less confusion for similar object categories
Finally, CDSSD achieves high-precision at high-recall range 
and outperforms YOLO and SSD (Tab.~\ref{tab:precision-recall}).

%\vspace{-0.5cm}

\subsection{Fashion Trends}
Using the object detection technique,
we could identify several fashion trends
on social media platform.
We mainly did analysis of Indian social media trends.
We could predict the use usage of palazzos
in 2015 before they became famous in India in 2016.
By parsing celebrity photos,
we could predict `green' as the dominant color
for hand-bags for 2018.
Parsing photos published by fashion weeks and fashion shows,
we could identify light gray color as the dominant
color for long dresses.
Furthermore, we could identify plum as the complementary
color for tops that are paired with black jeans.
This level of trend analysis would not have been
possible without deeply parsing photos
and extracting each fashion object along with its class
and bounding box.

\begin{figure*}
\centering
\includegraphics[height=15cm]{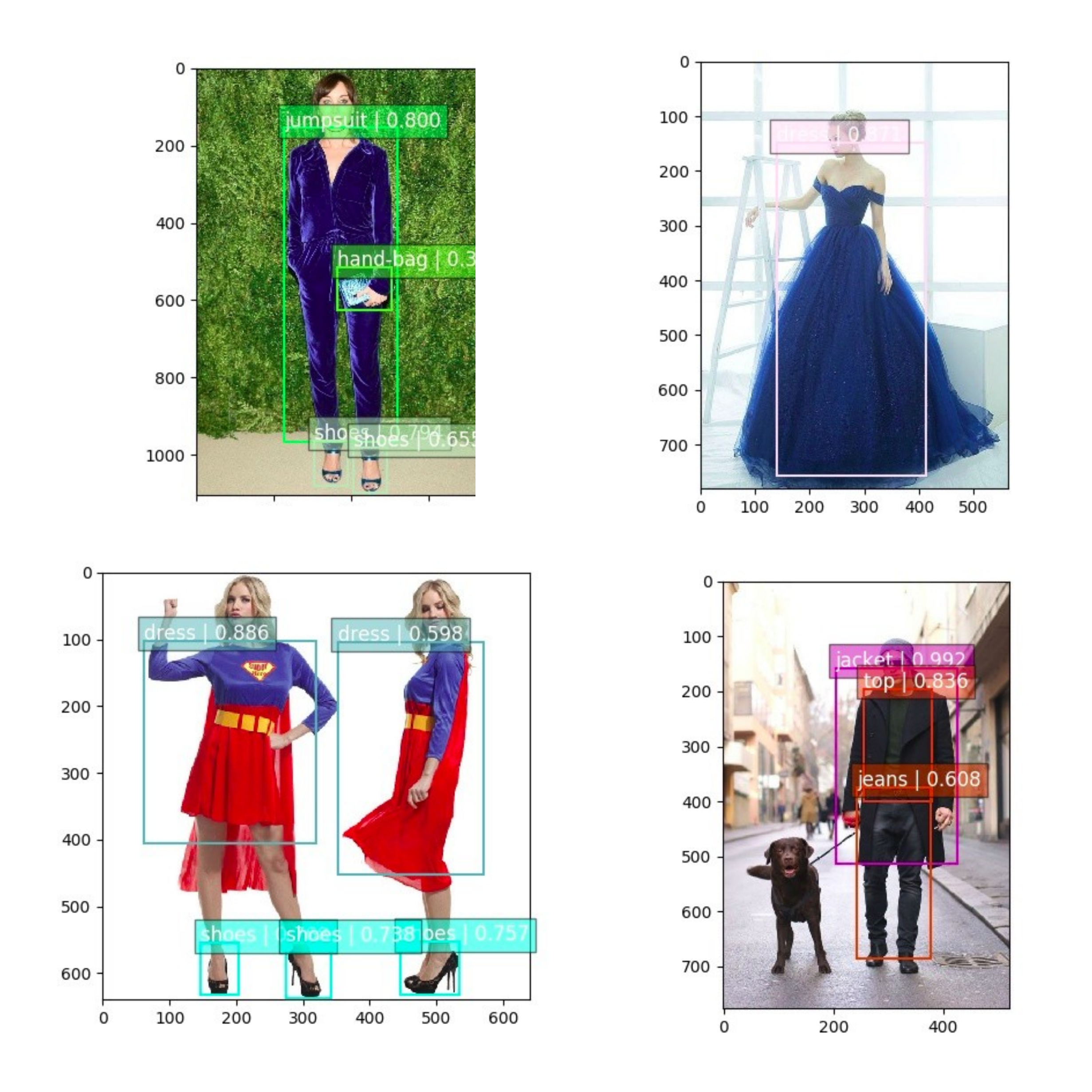}
\caption{Results of fashion detection}
\label{fig:ssd-stairnet-bssd}
\end{figure*}

\section{Conclusion}
We design an end-to-end 
framework using convolution-deconvolution
deep networks to improve the state-of-the-art
of single shot object detection techniques.
%Our approach combines feature maps from
%small and large receptive fields from
%both convolution
%and deconvolution blocks to yield a superior
%object detector.
%We experimentally demonstrate the effectiveness
%of our design choices.
Using a combination of unsupervised learning
and confluence of feature maps with different receptive fields, 
we demonstrate substantial improvement in mAP for different objects
in PASCAL VOC and MS COCO datasets
while reducing the bounding box requirement by 8 times,
thus improving inference time by 10\%.
%As a future work, our approach can be used to improve region proposal based
%detection techniques as well.
We believe that our work will inspire extensions
to region proposal based detection techniques
as well as other genres of objection detection
towards finding more effective and efficient ways to combine
feature maps of convolution and deconvolution blocks.
%to improve image classification, object detection
%and semantic segmentation approaches.

%\bibliographystyle{ACM-Reference-Format}
%\bibliography{biblio}

\end{document}